\documentclass[conference]{IEEEtran}
\IEEEoverridecommandlockouts
\usepackage{tabularx}
\usepackage{placeins}
\usepackage{cite}
\usepackage{amsmath,amssymb}
\usepackage{booktabs}
\usepackage{graphicx}
\usepackage{microtype}
\usepackage{multirow}
\usepackage{url}
\usepackage{xcolor}
\usepackage{balance}
\usepackage{etoolbox}
\usepackage{enumitem}
\usepackage{xurl}
\usepackage[hidelinks]{hyperref}

\newcommand{\pp}{\,\mathrm{pp}}

\begin{document}

\title{Sample Count Is Not Enough:\\
Candidate-Generation Strategy Shapes the Energy and Performance of LLM Test-Time Scaling}

\author{
\IEEEauthorblockN{Mobina Kashaniyan and Ali Jannesari}
\IEEEauthorblockA{
Department of Computer Science\\
Iowa State University, USA \\
\{mobina, jannesar\}@iastate.edu
}
}

\maketitle
\begin{abstract}
Test-time scaling can improve large language model reasoning by generating
and combining multiple candidate responses. In sampling-based methods, the
inference budget is often described by the number of generated candidates,
$N$. However, $N$ tells us how many candidates are generated, not how they
are executed. The same candidate budget can be produced in one batched
generation call or split across several sequential calls with smaller batch
sizes. We first study the effect of increasing $N$ on reasoning accuracy using
Phi-3-mini and Qwen2.5-1.5B on 500 GSM8K prompts. As expected, increasing
$N$ from 1 to 8 improves accuracy by 8.4 percentage points for Phi-3-mini
and 18.4 points for Qwen2.5-1.5B. However, accuracy alone does not show the
systems cost of using a larger candidate budget. We therefore fix $N=8$ and compare four generation schedules:
$1\times8$, $2\times4$, $4\times2$, and $8\times1$, where $a\times b$
denotes $a$ generation calls with $b$ candidates per call. We measure
latency, throughput, GPU-hours, and gross GPU-device energy while keeping
the total candidate count fixed. On A100 GPUs, eight serial calls use
$4.64$--$4.86\times$ as much gross GPU-device energy and have
$5.77$--$6.12\times$ the P95 latency of one batched call with eight
candidates. The same pattern appears across three independently scheduled
A100 nodes per model and in short-output SciQ/V100 experiments. These results show that candidate count alone is not enough to describe the
systems cost of multi-candidate test-time scaling. When candidates are
independent and memory allows it, fewer generation calls with larger batch
sizes are more efficient. Evaluations should therefore report not only
candidate count and accuracy, but also generation schedule and GPU-level
systems metrics.
\end{abstract}


\section{Introduction}
Test-time scaling improves large language model (LLM) reasoning by allocating
additional inference compute, often through multiple sampled responses.
Self-consistency, for example, generates several reasoning paths and selects
the most frequent final answer~\cite{wang2023selfconsistency}. In these
methods, the inference budget is often summarized by the candidate count,
$N$. However, $N$ tells us how many candidates are generated, not how they
are executed. The same candidate budget can be generated in one batched call
or split across several sequential calls with smaller batch sizes. Although
these schedules use the same number of candidates and the same aggregation
rule, they require different numbers of generation calls and use different
batch sizes. This can lead to large differences in latency, throughput,
GPU-hours, utilization, and energy.
We study this execution choice at $N=8$ using $1\times8$, $2\times4$,
$4\times2$, and $8\times1$ schedules. This issue is especially relevant in
HPC settings, where LLM inference may run as finite batch jobs rather than
continuous serving workloads. Candidate generation may also be divided across
calls for logging, deterministic seeding, control logic, or intermediate
analysis. Although serving research has shown that batching and scheduling
affect inference efficiency~\cite{yu2022orca,kwon2023vllm,
agrawal2024sarathi}, Table~\ref{tab:reporting-audit} shows that representative
test-time-scaling studies often report candidate budgets without reporting
calls per query, candidates per call, or measured energy. This makes systems
results harder to reproduce and compare.
We address three questions:
\begin{enumerate}
    \item How do accuracy and systems cost change as batched $N$ increases
    from 1 to 8?
    \item At fixed $N=8$, how does generation schedule affect latency,
    throughput, GPU-hours, utilization, and energy?
    \item Do these effects remain across different GPU nodes and in a
    short-output workload?
\end{enumerate}
This paper makes three contributions:
\begin{itemize}
    \item \textbf{Execution schedule and reporting gap.} We formalize the
    candidate-generation schedule as $S=(b_1,\ldots,b_C)$ and show that
    candidate count $N$ alone does not fully describe a multi-candidate
    inference workload. We also show that calls per query and candidates per
    call are often missing from representative test-time-scaling studies.
    \item \textbf{Fixed-budget systems characterization.} At fixed $N=8$,
    we measure the end-to-end effect of $1\times8$, $2\times4$,
    $4\times2$, and $8\times1$ schedules on latency, throughput, GPU-hours,
    utilization, and gross GPU-device energy. On A100 GPUs, serial execution
    uses $4.64$--$4.86\times$ as much energy as a single eight-candidate
    batched call.
    \item \textbf{Robustness and practical guidance.} We evaluate additional
    A100 nodes, two models, and a short-output SciQ/V100 case study. The
    results support a practical guideline: when candidates are independent
    and memory allows it, fewer generation calls with larger batch sizes are
    more efficient. We also provide minimum reporting recommendations for
    multi-candidate inference experiments.
\end{itemize}
These results show that the system cost of test-time scaling depends not only
on how many candidates are generated, but also on how those candidates are
grouped into generation calls.

\section{Background and Related Work}

\subsection{Test-Time Scaling}

Test-time scaling improves LLM outputs by using additional compute during
inference. Some methods spend this compute on extending or refining a
reasoning trajectory, while others generate and combine multiple candidate
responses. Prior work shows that additional test-time compute can improve
reasoning and can sometimes help smaller models approach the performance of
larger ones under similar inference budgets
~\cite{snell2024scaling,muennighoff2025s1,zhang2025ttssurvey}. However, the
benefit depends on factors such as prompt difficulty, reasoning strategy,
stopping, and aggregation~\cite{ghosal2025mirage}. Our work focuses on
another factor: how a multi-candidate inference budget is executed on the
underlying hardware.

\subsection{Multi-Candidate Sampling}

Self-consistency samples multiple reasoning paths and selects the most
frequent answer~\cite{wang2023selfconsistency}, while best-of-$N$ selects a
candidate using a verifier, reward model, or confidence measure
~\cite{kang2025selfcertainty}. Prior studies examine larger or adaptive
sampling budgets, stopping rules, and different selection methods
~\cite{aggarwal2023adaptive,chen2023universal,chen2024morecalls,
wang2025dasc,ding2025bestroute,kim2026reasc,iwase2026prefix,
kashaniyan2026interpretableadaptivesamplingllm}. These works mainly study how
many candidates to generate or how to select among them. In contrast, we keep
the candidate budget fixed and study how executing the same candidates across
different numbers of generation calls and batch sizes affects systems cost.

\subsection{Inference Scheduling and Energy}

LLM inference efficiency depends on batching, scheduling, memory management,
and resource allocation~\cite{li2024llmserving}. ORCA, vLLM, and
Sarathi-Serve improve utilization and throughput through different batching
and scheduling strategies
~\cite{yu2022orca,kwon2023vllm,agrawal2024sarathi}. Other work studies
heterogeneous scheduling and KV-cache constraints
~\cite{yang2024perllm,jaillet2025online}. Energy consumption also depends on
model size, hardware, batch size, sequence length, parallelism, and
utilization~\cite{ding2024sustainable,stojkovic2024greener,jegham2025hungry,
ozcan2025quantifying,wilkins2025offlineenergy}.
Recent work has also compared the performance and energy efficiency of LLM
inference across different AI accelerators and batch sizes
~\cite{brunetta2026beyond}. These studies show that execution choices can
strongly affect inference cost. Our work connects these systems factors to
multi-candidate test-time scaling by measuring how generation-call structure
affects latency, GPU-hours, utilization, and gross GPU-device energy at fixed
$N$.

\subsection{Reporting Practices in Prior Work}

Table~\ref{tab:reporting-audit} audits representative foundational,
adaptive-budget, and repeated-sampling studies. We record a field as reported
only when the corresponding execution detail is stated explicitly in the
paper or its supplementary material. ``NR'' denotes not reported. The audit
is intended to characterize reporting practices in representative work rather
than provide an exhaustive systematic review. Representative studies commonly
report candidate budgets but do not fully specify generation-call structure
or measured energy cost.

\begin{table}[t]
\centering
\caption{Reporting practices in representative multi-candidate and
test-time-scaling studies. HW refers to candidate-generation hardware;
``NR'' means not reported.}
\label{tab:reporting-audit}
\footnotesize
\renewcommand{\arraystretch}{1.08}
\setlength{\tabcolsep}{2.3pt}

\begin{tabularx}{\columnwidth}{@{}Xccccc@{}}
\toprule
Study
& $N$
& \shortstack{Calls/\\query}
& \shortstack{Cand./\\call}
& HW
& Energy \\
\midrule

Self-Consistency~\cite{wang2023selfconsistency}
& Yes & NR & NR & NR & NR \\

Adaptive-Consistency~\cite{aggarwal2023adaptive}
& Yes & NR & NR & NR & NR \\

Universal Self-Consistency~\cite{chen2023universal}
& Yes & NR & NR & NR & NR \\

Large Language Monkeys~\cite{brown2024monkeys}
& Yes & NR & NR & Yes & NR \\

Scaling Test-Time Compute~\cite{snell2024scaling}
& Yes & NR & NR & Yes & NR \\

Difficulty-Adaptive SC~\cite{wang2025dasc}
& Yes & NR & NR & NR & NR \\

\midrule
\textbf{This work}
& \textbf{Yes}
& \textbf{Yes}
& \textbf{Yes}
& \textbf{Yes}
& \textbf{Yes} \\

\bottomrule
\end{tabularx}
\end{table}

This work reports $N$, calls per query, candidates per call, inference
hardware, and gross GPU-device energy.

\section{Methods}

\subsection{Candidate-Generation Strategies}

\begin{figure}[!ht]
    \centering
    \includegraphics[width=\columnwidth]{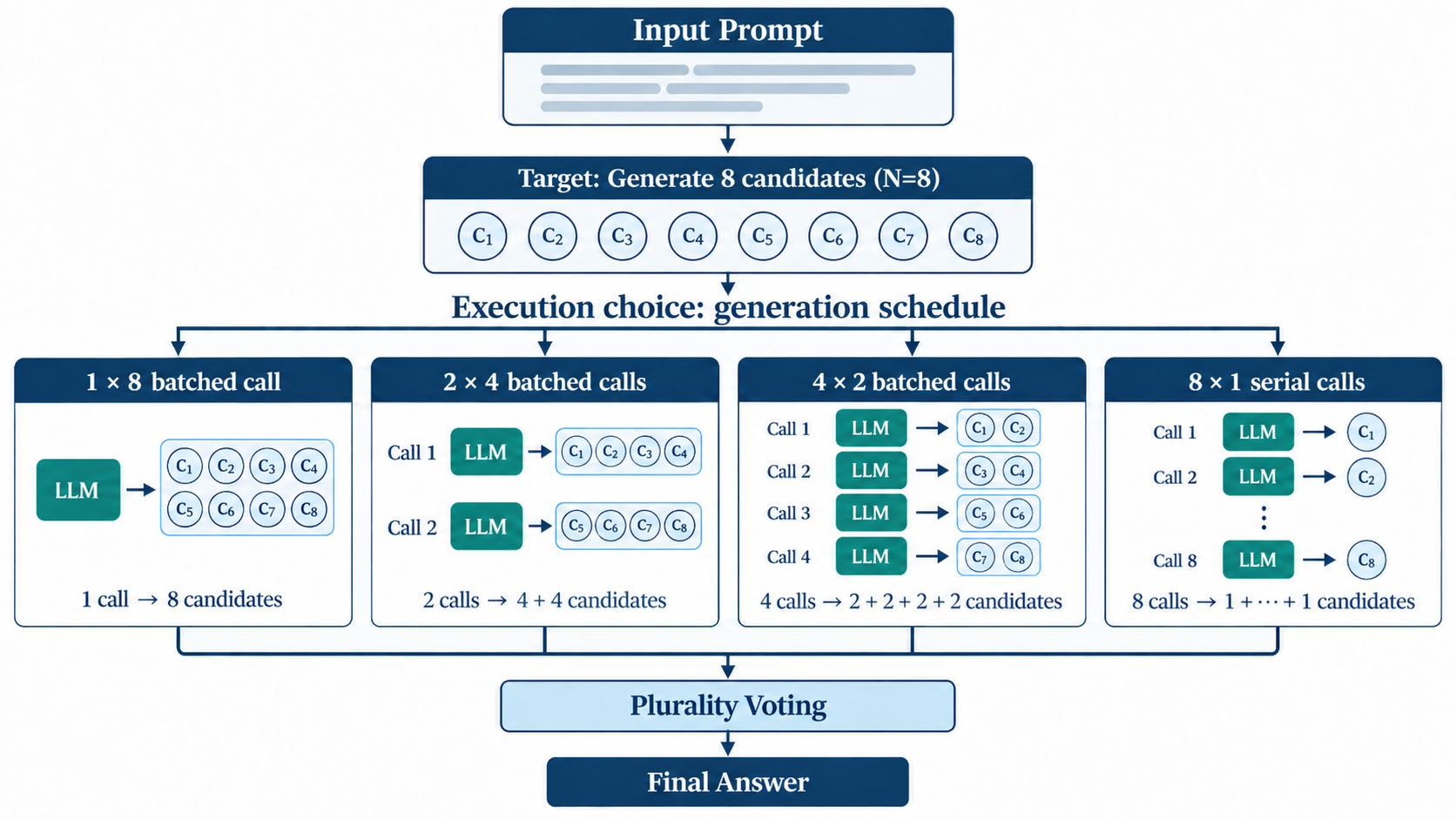}
    \caption{Generation schedules for a fixed candidate budget of $N=8$.
    All schedules use the same prompt, generate eight candidates, and apply
    the same plurality vote. They differ in the number of sequential
    generation calls and candidates generated per call. Repeated LLM blocks
    represent successive calls on the same GPU.}
    \label{fig:generation-schedules}
\end{figure}

Our goal is to separate how many candidates are generated from how those
candidates are executed. Let $N$ be the total number of candidates generated
for a prompt. We represent the generation schedule as
\[
\mathcal{S}=(b_1,\ldots,b_C), \qquad
\sum_{c=1}^{C} b_c=N,
\]
where $C$ is the number of generation calls and $b_c$ is the number of
candidates generated in call $c$.
Figure~\ref{fig:generation-schedules} shows the workflow for $N=8$. We compare
$1\times8$, $2\times4$, $4\times2$, and $8\times1$ schedules. In an
$a\times b$ schedule, $a$ is the number of generation calls and $b$ is the
number of candidates generated in each call. The calls are executed
sequentially on the same allocated GPU, while candidates within each call are
generated together as a batch. The four schedules therefore correspond to
$\mathcal{S}=(8)$, $(4,4)$, $(2,2,2,2)$, and
$(1,1,1,1,1,1,1,1)$.
All schedules use the same prompts, decoding settings, answer extraction, and
voting procedure. Candidate responses are sampled independently across
schedules in the systems experiments. We also study how accuracy changes as the candidate budget increases. For
each prompt, we generate eight candidates and compute accuracy for
$N\in\{1,2,3,4,8\}$ using the first $N$ candidates. This gives paired
comparisons across candidate counts. These candidates are used only for the
accuracy analysis. Systems measurements are collected separately by running
each schedule on the GPU.

\subsection{Token Accounting}

We track logical token volume to check that differences between schedules are
not caused by large differences in generated response length. Let $P_q$ be
the prompt length for query $q$. Since each candidate is generated from the
same prompt, the candidate-associated logical input volume is
\[
T_{\mathrm{input}}^{\mathrm{logical}}(q)=NP_q.
\]
For fixed $N=8$, this is $8P_q$ for every schedule.
Let $L_{q,c,j}$ be the generated length of candidate $j$ in call $c$. The
logical number of generated tokens is
\[
T_{\mathrm{gen}}^{\mathrm{logical}}(q,\mathcal{S})
=
\sum_{c=1}^{C}\sum_{j=1}^{b_c}L_{q,c,j}.
\]
In the fixed-$N$ experiments, the candidate count is identical across
schedules and the logical generated-token volume remains closely matched.
This lets us compare the systems cost of different generation schedules while
keeping the overall candidate budget fixed.
\subsection{Answer Extraction and Voting}
After generation, all candidates use the same answer-extraction and
plurality-voting procedure. Candidates with the same extracted answer form a
group, and the largest group determines the final prediction. We do not use a
verifier, reward model, or token-level score. For GSM8K, we extract the numeric final answer by checking the requested
final-answer delimiter, boxed expressions, explicit answer statements,
trailing numeric expressions, and the last non-empty line. Numeric answers
are then normalized to a common form. For SciQ, we extract one of the answer
choices A--D case-insensitively. Extraction failures remain possible voting outcomes and are counted as
incorrect if selected. If multiple answers receive the same number of votes,
we select the answer that appears first among the generated candidates. In
the accuracy analysis, this rule can make $N=2$ identical to $N=1$ when the
first two candidates disagree. We therefore also evaluate $N=3$ and report a sensitivity analysis using uniform random tie-breaking.
\subsection{Measurement Boundary and Energy}
Each measured query begins with GPU synchronization and an initial NVML
cumulative-energy reading. The measured interval includes prompt processing,
prefill, decoding, all generation calls, answer extraction, and plurality
voting. After the final vote, we synchronize the GPU again and record the
final cumulative-energy value. Model loading, warm-up, reporting-time token
counting, and final grading are excluded.
Gross GPU-device energy is computed as
\[
E_{\mathrm{gross}}
=
E_{\mathrm{NVML,end}}-E_{\mathrm{NVML,start}}.
\]
Idle energy is not subtracted. The reported values therefore represent gross
GPU-device energy over the full measured query interval, not isolated dynamic
computation energy or whole-node energy. Average GPU power is computed as
gross energy divided by query latency. A schedule can therefore have lower
average power but still consume more total energy if it keeps the GPU active
for longer.
\section{Experimental Design and Measurement}

\subsection{Study Overview}

We organize the evaluation around three goals. First, we measure how accuracy
changes as the candidate budget increases. Second, we measure how systems
cost changes with candidate count and with generation schedule. Third, we
check whether the main scheduling effect remains across different GPU nodes
and on a short-output workload. Table~\ref{tab:study-matrix} summarizes the
five experiments used for these goals.
\begin{table}[!ht]
\centering
\caption{Summary of the experimental studies.}
\label{tab:study-matrix}
\footnotesize
\renewcommand{\arraystretch}{1.08}
\setlength{\tabcolsep}{2.5pt}
\begin{tabularx}{\columnwidth}{@{}lXX@{}}
\toprule
Study & Workload & Design \\
\midrule

Accuracy scaling
& GSM8K, 500 prompts
& One eight-candidate pool; evaluate
$N\in\{1,2,3,4,8\}$ using prefixes \\

Batched scaling
& GSM8K, $100\times3$
& Batched $N\in\{1,2,4,8\}$ and serial $N=8$ \\

Schedule sweep
& GSM8K, $100\times3$
& Fixed $N=8$: $1\times8$, $2\times4$,
$4\times2$, $8\times1$ \\

Cross-node check
& GSM8K, $100\times3$
& Repeat $1\times8$ and $8\times1$ on two
additional A100 nodes \\

Short-output validation
& SciQ, $500\times3$
& Complete fixed-$N=8$ sweep on V100 GPUs \\

\bottomrule
\end{tabularx}
\end{table}
For the accuracy study, each model generates eight candidates for 500 GSM8K
prompts. Accuracy for smaller values of $N$ is computed using the first $N$
candidates from the same set. This gives paired prompt-level comparisons
across candidate budgets. These generations are used only for accuracy
analysis; their latency and energy are not assigned to individual values of
$N$. The GSM8K systems experiments use the same 100 prompts over three
repetitions. The batched-scaling study measures how systems cost changes as
$N$ increases. The main schedule sweep instead fixes $N=8$ and changes only
the number of generation calls and candidates per call. All four schedules
are executed within the same A100 job. Two repetitions use the order
$1\times8$, $2\times4$, $4\times2$, $8\times1$, while one uses the reverse
order to reduce possible order and thermal effects. To test run-to-run stability, we repeat the $1\times8$ and $8\times1$
endpoints on two additional A100 nodes per model. Together with the primary
job, this gives three independently scheduled A100 jobs per model. Finally, we repeat the full fixed-$N=8$ schedule sweep on 500 SciQ prompts
over three repetitions using V100 GPUs. SciQ produces much shorter responses
than GSM8K, so we use it as a separate short-output validation. It is not
intended as a complete study of how output length affects scheduling cost.
\subsection{Prompts, Seeds, and Warm-Up}
The GSM8K test split is shuffled with seed 42. The first 100 prompts are used
for the systems experiments and are also part of the 500-prompt accuracy
study. Prompt order is fixed across schedules, repetitions, jobs, and nodes. We use deterministic method-specific seeds so that each run is reproducible
while schedules sample candidates independently. Two warm-up generations are
performed after model loading and are excluded from measurement.
\subsection{Models, Workloads, and Hardware}
We evaluate Phi-3-mini-4k-instruct~\cite{abdin2024phi3} and
Qwen2.5-1.5B-Instruct~\cite{yang2024qwen25} on GSM8K
~\cite{cobbe2021gsm8k} and SciQ~\cite{welbl2017sciq}. Sampling is enabled
with temperature 1.0 and top-$p$ 0.95.
\begin{table}[!ht]
\centering
\caption{GSM8K batched-scaling configurations.}
\label{tab:setup}
\footnotesize
\renewcommand{\arraystretch}{1.08}
\setlength{\tabcolsep}{3pt}

\begin{tabularx}{\columnwidth}{@{}Xcc@{}}
\toprule
Configuration & Phi-3 & Qwen \\
\midrule
GPU & V100 32\,GB & A100 80\,GB \\
Power limit & 250\,W & 500\,W \\
Driver / CUDA & 580.159.04 / 12.1 & 580.159.04 / 12.1 \\
PyTorch / Transformers & 2.4.1 / 4.57.6 & 2.4.1 / 4.57.6 \\
Prompts $\times$ repetitions & $100\times3$ & $100\times3$ \\
Maximum new tokens & 512 & 512 \\
\bottomrule
\end{tabularx}
\end{table}
The batched-scaling experiment runs Phi-3 on a V100 and Qwen on an A100,
we interpret each model--GPU pair separately rather than compare their
absolute systems values. The primary schedule and cross-node experiments run
both models on A100-SXM4 80\,GB GPUs with a 500\,W power limit and a fixed
1275\,MHz graphics clock. The SciQ experiments use V100 PCIe 32\,GB GPUs
with a 250\,W power limit and a fixed 1230\,MHz graphics clock. Each job
reserves one GPU exclusively.
\subsection{Output-Length Validation}
GSM8K uses a maximum output length of 512 tokens. In the 500-prompt accuracy
sets, 2.0\% of Phi-3 candidates and 5.25\% of Qwen candidates reach this
limit, with mean response lengths of 251 and 270 tokens, respectively. SciQ uses a 64-token limit. No Qwen candidates and 1.93\% of Phi-3 candidates
reach the limit, with mean response lengths of 1.71 and 3.46 tokens,
respectively. These values confirm that SciQ provides a much shorter-output
workload than GSM8K.
\subsection{Statistical Analysis}
Systems results are reported as mean$\pm$SD across three repetitions. P95
latency is computed within each repetition and then summarized across
repetitions. Accuracy confidence intervals use 10{,}000 prompt-level bootstrap resamples,
with paired resampling when comparing candidate budgets. For the fixed-$N$
schedule study, ratios compare $8\times1$ with $1\times8$. Confidence
intervals use a hierarchical paired bootstrap: repetitions are resampled
first, followed by prompts within each selected repetition, while preserving
the pairing between schedules. P95 latency is recomputed for each bootstrap
sample. These confidence intervals describe variability within the primary jobs.
The two additional A100 jobs are analyzed separately, and cross-node results
are reported as the observed range across the three jobs.

\section{Results}

\subsection{Accuracy Gains from Increasing $N$}
\label{sec:accuracy}

\begin{figure}[!ht]
\centering
\includegraphics[width=\columnwidth]
{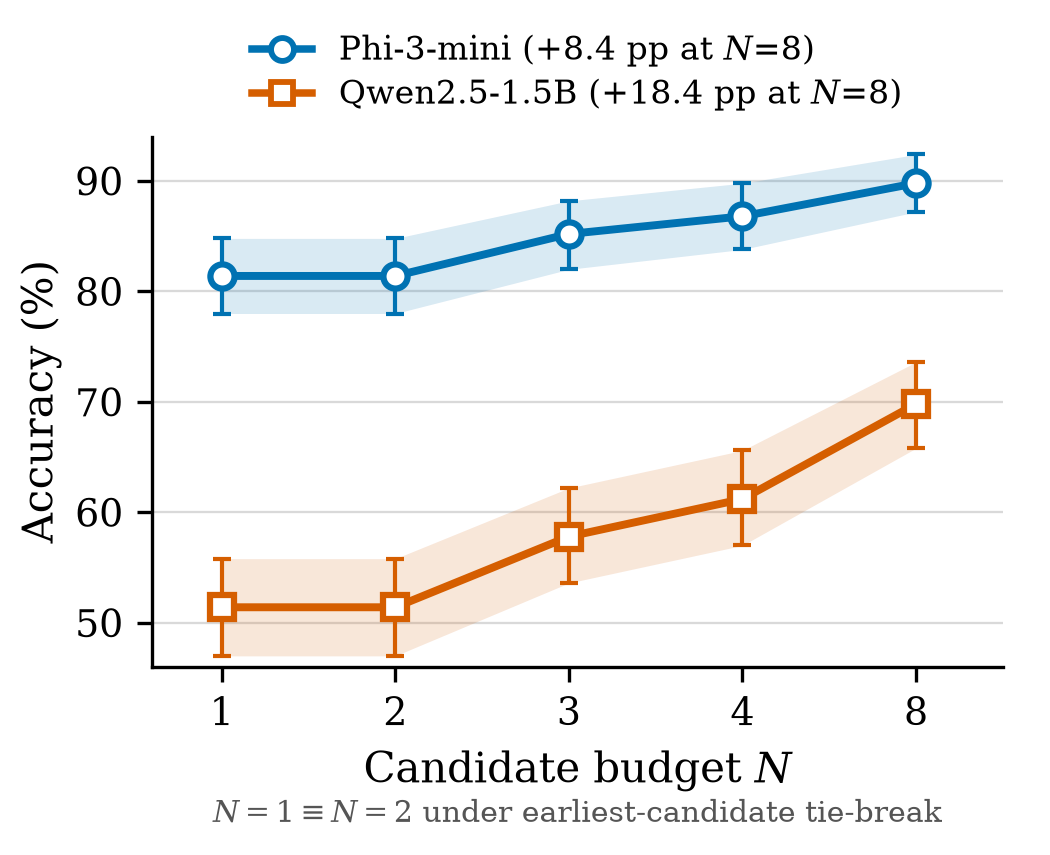}
\caption{GSM8K accuracy as the candidate budget increases. Error bars show
95\% bootstrap intervals over 500 prompts.}
\label{fig:accuracy}
\end{figure}

Figure~\ref{fig:accuracy} shows how GSM8K accuracy changes as the candidate
budget increases. As expected, generating more candidates improves accuracy.
Phi-3 increases from 81.4\% at $N=1$ to 89.8\% at $N=8$, a gain of
$8.4\pp$ with a 95\% paired bootstrap interval of $[5.8,11.2]\pp$.
Qwen increases from 51.4\% to 69.8\%, a gain of $18.4\pp$
($[14.8,22.0]\pp$).

The identical accuracy at $N=1$ and $N=2$ comes from the tie-breaking rule.
When the first two candidates disagree, each receives one vote and the first
candidate is selected. Accuracy begins to increase at $N=3$, when a majority
can form.

Table~\ref{tab:ties} checks whether tie handling or answer extraction explains
the observed gains. Random tie-breaking changes expected accuracy by at most
$1.4\pp$, while conditioning on successful extraction also produces only
small changes. We therefore use the original plurality-voting result as the
main accuracy measure.

\begin{table}[!ht]
\centering
\caption{Tie and extraction diagnostics. ``Random'' uses uniform random
selection among tied groups. ``Cond.'' conditions on a successfully
extracted selected answer.}
\label{tab:ties}
\scriptsize
\renewcommand{\arraystretch}{1.08}
\setlength{\tabcolsep}{2pt}
\begin{tabularx}{\columnwidth}{@{}Xccccc@{}}
\toprule
Model & $N$ & Tie (\%) & Primary & Random & Cond. \\
\midrule
\multirow{3}{*}{Phi-3}
 & 2 & 25.0 & 81.4 & 81.9 & 82.1 \\
 & 4 & 7.0 & 86.8 & 87.3 & 86.8 \\
 & 8 & 3.4 & 89.8 & 90.1 & 89.8 \\
\midrule
\multirow{3}{*}{Qwen}
 & 2 & 63.0 & 51.4 & 50.5 & 52.1 \\
 & 4 & 31.4 & 61.2 & 61.3 & 61.8 \\
 & 8 & 18.2 & 69.8 & 71.2 & 70.5 \\
\bottomrule
\end{tabularx}
\end{table}

\FloatBarrier
\subsection{Systems Cost of Increasing Batched $N$}

We next measure what happens to systems cost as the batched candidate budget
increases from $N=1$ to $N=8$. Phi-3 and Qwen use different GPUs in this
experiment, so their absolute values are not compared directly.

Table~\ref{tab:batched-energy} highlights the main energy result. For both
models, energy per query increases as more candidates are generated, while
energy per generated token decreases. For Phi-3, energy increases from
631 to 1286\,J/query, while energy per token decreases from 2.634 to
0.655\,J. For Qwen, the corresponding values change from 596 to
975\,J/query and from 2.222 to 0.459\,J/token.

\begin{table}[!ht]
\centering
\caption{Energy cost as the batched candidate budget increases. Values are
mean$\pm$SD across three repetitions. Phi-3 uses a V100 and Qwen uses an
A100, so absolute values are not compared across models.}
\label{tab:batched-energy}
\scriptsize
\renewcommand{\arraystretch}{1.08}
\setlength{\tabcolsep}{2pt}
\begin{tabularx}{\columnwidth}{@{}cXXXX@{}}
\toprule
$N$ &
\multicolumn{2}{c}{Phi-3/V100} &
\multicolumn{2}{c}{Qwen/A100} \\
\cmidrule(lr){2-3}\cmidrule(lr){4-5}
& J/query & J/token & J/query & J/token \\
\midrule
1 & $631\pm19$ & $2.634\pm0.005$ & $596\pm13$ & $2.222\pm0.025$ \\
2 & $795\pm15$ & $1.607\pm0.005$ & $670\pm19$ & $1.265\pm0.013$ \\
4 & $974\pm36$ & $1.004\pm0.025$ & $802\pm6$  & $0.758\pm0.007$ \\
8 & $1286\pm49$ & $0.655\pm0.019$ & $975\pm20$ & $0.459\pm0.011$ \\
\bottomrule
\end{tabularx}
\end{table}

Larger batches also increase token throughput, but total query latency still
rises. From $N=1$ to $N=8$, mean latency increases from 5.06 to 8.91\,s for
Phi-3 and from 5.26 to 7.64\,s for Qwen. Measured GPU-hours per 1{,}000
queries also increase from 1.41 to 2.47 and from 1.46 to 2.12,
respectively. Thus, batching improves per-token efficiency, but generating
more candidates still increases the total cost of a query.

\FloatBarrier
\subsection{Effect of Generation Schedule at Fixed $N=8$}

The previous experiment changes the candidate count. We now keep the
candidate budget fixed at eight and change only how those candidates are
grouped into generation calls. Mean logical generated-token volume varies by
only 0.8\% across Phi-3 schedules and 1.0\% across Qwen schedules.

Table~\ref{tab:schedule-trend} shows a clear trend. Splitting the same eight
candidates across more calls increases both energy and P95 latency for both
models. The intermediate schedules follow the same pattern, showing that the
effect is gradual rather than appearing only at the fully serial endpoint.

\begin{table}[!ht]
\centering
\caption{Fixed-$N=8$ schedule sweep. Values are normalized to the
$1\times8$ schedule. Lower values are better.}
\label{tab:schedule-trend}
\footnotesize
\renewcommand{\arraystretch}{1.08}
\setlength{\tabcolsep}{2.2pt}
\begin{tabularx}{\columnwidth}{@{}Xcccc@{}}
\toprule
& \multicolumn{2}{c}{Relative energy} &
\multicolumn{2}{c}{Relative P95 latency} \\
\cmidrule(lr){2-3}\cmidrule(lr){4-5}
Schedule & Phi-3 & Qwen & Phi-3 & Qwen \\
\midrule
$1\times8$ & 1.00 & 1.00 & 1.00 & 1.00 \\
$2\times4$ & 1.63 & 1.66 & 1.81 & 1.97 \\
$4\times2$ & 2.71 & 2.86 & 3.21 & 3.57 \\
$8\times1$ & 4.64 & 4.86 & 5.77 & 6.12 \\
\bottomrule
\end{tabularx}
\end{table}

At the fully serial endpoint, eight single-candidate calls use
$4.64\times$ as much gross GPU-device energy as one eight-candidate call
for Phi-3 (95\% CI: $[4.48,4.79]$) and $4.86\times$ as much for Qwen
($[4.71,5.02]$). P95 latency reaches $5.77\times$
($[5.38,5.99]$) and $6.12\times$ ($[5.70,6.75]$), while throughput falls
to 16.7\% and 17.9\% of the batched baseline.

The practical cost is also visible in GPU time. For 1{,}000 queries, measured
GPU time increases from 2.09 to 12.49 GPU-hours for Phi-3 and from 2.13 to
11.76 GPU-hours for Qwen. Lower average power does not remove this penalty.
For example, Phi-3 mean power decreases from 177.8\,W to 139.8\,W, but mean
latency increases by $5.97\times$, so total energy still increases.

These results give a simple practical guideline for the settings studied
here: when candidates are independent and memory allows it, fewer generation
calls with larger batch sizes are more efficient.

\FloatBarrier
\subsection{Cross-Node Robustness}
\label{sec:crossnode}

We repeat the $1\times8$ and $8\times1$ endpoints in three independently
scheduled A100 jobs per model. Table~\ref{tab:crossnode} shows that the main
effect remains stable across these jobs.

\begin{table}[!ht]
\centering
\caption{Observed $8\times1$/$1\times8$ ratio ranges across three
independent A100 jobs per model.}
\label{tab:crossnode}
\footnotesize
\renewcommand{\arraystretch}{1.08}
\setlength{\tabcolsep}{3pt}
\begin{tabularx}{\columnwidth}{@{}Xcc@{}}
\toprule
Metric & Phi-3 & Qwen \\
\midrule
Gross J/query $\downarrow$ & 4.43--4.64 & 4.85--4.88 \\
Mean latency $\downarrow$ & 5.85--5.97 & 5.50--5.53 \\
P95 latency $\downarrow$ & 5.53--5.77 & 6.06--6.12 \\
Throughput retained $\uparrow$ & 16.7--17.1\% & 17.9--18.0\% \\
\bottomrule
\end{tabularx}
\end{table}

The observed ranges are narrow compared with the size of the scheduling
effect. Serial throughput remains about 17--18\% of batched throughput in
every job. The similar results across the three A100 jobs show that the scheduling
effect is consistent across the tested nodes. However, we do not assume that
the exact ratios will remain the same on other GPU architectures, clusters,
or software environments.

\FloatBarrier
\subsection{Length-Stratified GSM8K Analysis}
\label{sec:length_sensitivity}

We next check whether the scheduling effect appears only for prompts that
produce long responses. The 100 GSM8K systems prompts are divided into four
groups using mean candidate length from the separate accuracy generation.

\begin{figure}[!ht]
\centering
\includegraphics[width=\columnwidth]
{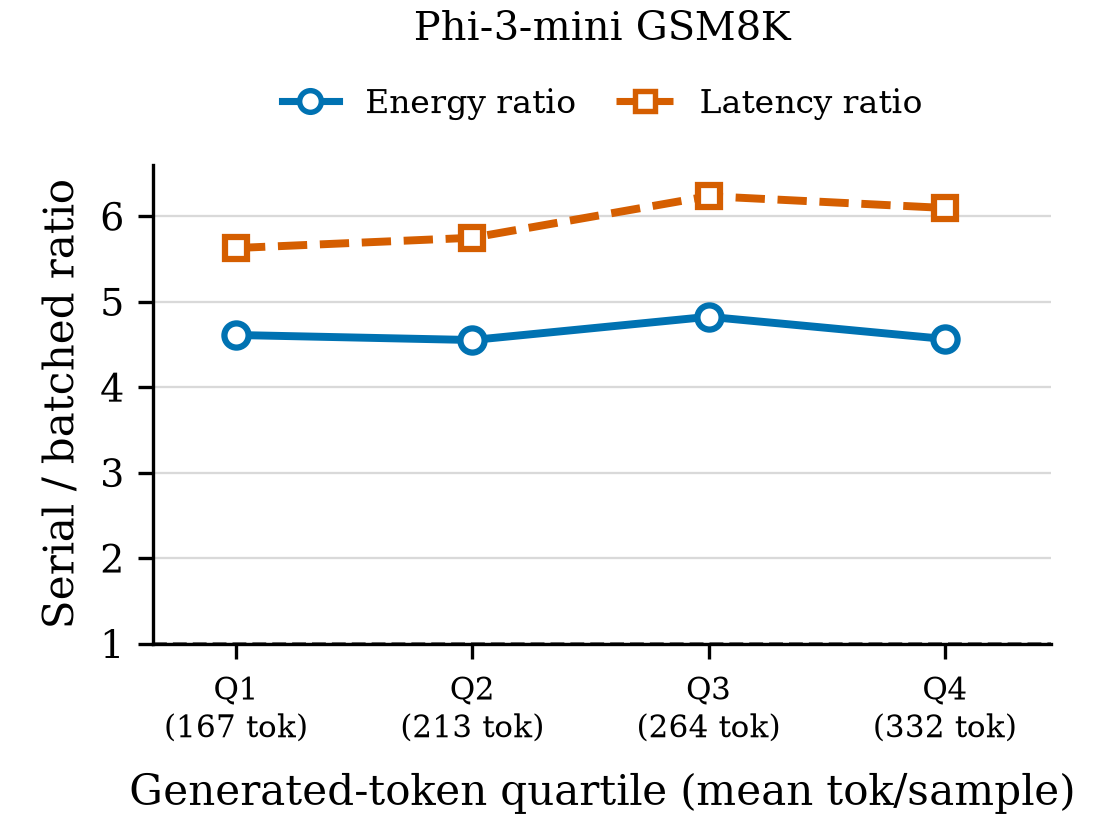}
\caption{Phi-3 serial-to-batched energy and latency ratios across GSM8K
response-length groups. Each group contains 25 prompts.}
\label{fig:length_phi3}
\end{figure}

For Phi-3, Figure~\ref{fig:length_phi3} shows energy ratios between
$4.40$ and $4.77$ and latency ratios between $5.45$ and $6.16$.
The ratios are not monotonic with response length.

\begin{figure}[!ht]
\centering
\includegraphics[width=\columnwidth]
{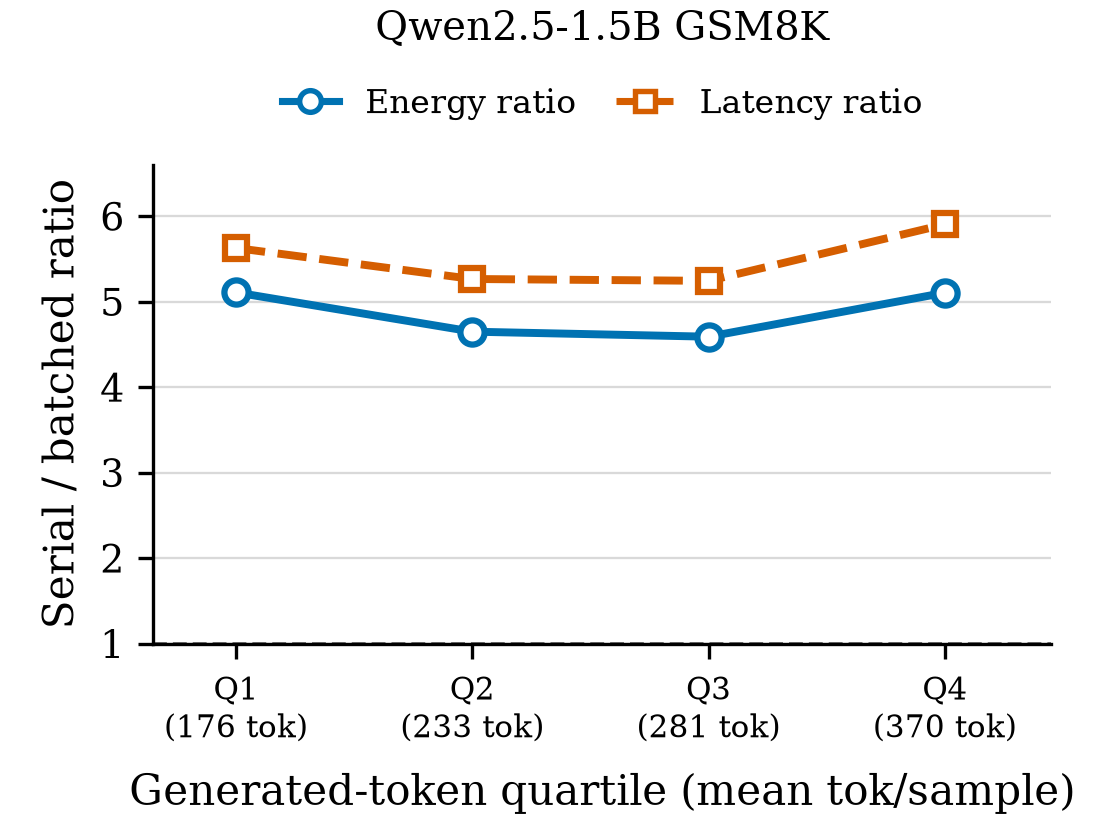}
\caption{Qwen serial-to-batched energy and latency ratios across GSM8K
response-length groups. Each group contains 25 prompts.}
\label{fig:length_qwen}
\end{figure}

Qwen shows the same general behavior. Figure~\ref{fig:length_qwen} shows
energy ratios between $4.51$ and $5.09$ and latency ratios between
$5.15$ and $5.88$.

The scheduling penalty therefore appears across all four response-length
groups rather than only for the longest outputs. This analysis is
descriptive because each group contains only 25 prompts and the serial and
batched schedules independently sample their candidates.

\FloatBarrier
\subsection{Short-Output Validation on SciQ}
\label{sec:sciq}

GSM8K still contains reasoning-style outputs even in its shortest group. We
therefore repeat the complete fixed-$N=8$ schedule sweep on SciQ, where mean
responses contain only a few generated tokens.

Table~\ref{tab:sciq-relative} shows the relative systems cost. The same trend
appears for both models: energy and latency increase as the candidate budget
is divided across more calls, while throughput decreases.
\begin{table}[!ht]
\centering
\caption{SciQ fixed-$N=8$ schedule sweep on V100 GPUs. Values are normalized
to $1\times8$ within each model.}
\label{tab:sciq-relative}
\footnotesize
\renewcommand{\arraystretch}{1.08}
\setlength{\tabcolsep}{2.3pt}
\begin{tabularx}{\columnwidth}{@{}Xcccc@{}}
\toprule
Model & Schedule & Rel. energy & Rel. latency & Rel. throughput \\
\midrule
\multirow{4}{*}{Phi-3}
& $1\times8$ & 1.00 & 1.00 & 1.00 \\
& $2\times4$ & 1.29 & 1.37 & 0.74 \\
& $4\times2$ & 1.75 & 1.97 & 0.51 \\
& $8\times1$ & 2.57 & 2.88 & 0.36 \\
\midrule
\multirow{4}{*}{Qwen}
& $1\times8$ & 1.00 & 1.00 & 1.00 \\
& $2\times4$ & 1.38 & 1.59 & 0.63 \\
& $4\times2$ & 2.04 & 2.59 & 0.39 \\
& $8\times1$ & 3.34 & 4.42 & 0.23 \\
\bottomrule
\end{tabularx}
\end{table}
From $1\times8$ to $8\times1$, gross energy increases by $2.57\times$ for
Phi-3 and $3.34\times$ for Qwen. Mean latency increases by $2.88\times$ and
$4.42\times$, while throughput falls to 36\% and 23\% of the batched
baseline. Because SciQ uses V100 GPUs while the primary GSM8K study uses A100 GPUs, we
do not claim that the difference in ratio size is caused only by output
length. The supported conclusion is narrower: the fixed-$N$ scheduling
effect appears in both the longer-output GSM8K/A100 setting and the
short-output SciQ/V100 setting.
\section{Discussion}

Our results show that candidate count alone does not fully describe the
systems cost of multi-candidate inference. For the same $N=8$ budget,
changing how candidates are grouped into generation calls produces large
differences in latency, throughput, GPU-hours, and gross GPU-device energy.
Across both models, dividing the candidate budget across more calls
consistently increases systems cost. The same pattern remains across
independently scheduled A100 jobs and also appears in the short-output
SciQ/V100 setting.
These results have a direct practical implication: when candidates are
independent and memory allows it, fewer generation calls with larger batches
are more efficient. This does not mean that $1\times N$ is always optimal.
Memory limits, request dependencies, continuous batching, and serving
constraints may require a different schedule.

The effect can become substantial at scale. Under the configurations studied
here, applying the measured per-query difference to one million queries would
add approximately 1.37\,MWh of gross GPU-device energy and 10{,}401 measured
GPU-hours for Phi-3 when moving from $1\times8$ to $8\times1$. For Qwen, the
corresponding differences are approximately 1.00\,MWh and 9{,}631 GPU-hours.
These values are linear illustrations of the measured configurations, not
projections to other deployments.

Multi-candidate evaluations should therefore report not only $N$, but also
the number of generation calls, candidates per call, batching mode, latency,
throughput, GPU-hours, and the energy-measurement boundary. Two experiments
with the same model, dataset, decoding settings, and candidate count can
otherwise have substantially different systems costs simply because their
candidate-generation schedules differ. Reporting these details makes systems
results easier to reproduce and compare.

\subsection{Why Generation Schedule Changes Systems Cost}

Keeping $N$ fixed keeps the candidate budget unchanged, but it does not keep
the execution pattern unchanged. In a larger batch, several candidates can
make progress within the same generation call. When the same candidates are
split across smaller calls, less work is exposed to the GPU at the same time,
and the query must pass through more generation calls before completion. Several factors can contribute to the additional cost. More calls can add
per-call framework and synchronization overhead and can repeat work associated
with processing the prompt and starting generation. Smaller calls can also
provide less parallel work to the GPU. Together, these effects can increase
execution time even though the total number of requested candidates remains
the same. The energy results also show why average power alone is not enough to judge
efficiency. A smaller or more serial workload may draw less power at a given
moment, but it can keep the GPU active for much longer. In our Phi-3/A100
measurements, mean power decreases from 177.8\,W for $1\times8$ to
139.8\,W for $8\times1$, while mean latency increases by $5.97\times$.
The longer execution time outweighs the lower average power, leading to
substantially higher total energy. These measurements capture the combined end-to-end effect of the generation
schedule. They do not isolate how much of the difference comes from repeated
prompt processing, call overhead, synchronization, GPU utilization, or other
low-level effects. Separating these mechanisms would require more detailed
profiling and is an important direction for future work.

\subsection{Practical Schedule Selection}

Our results suggest that candidate generation can be treated as a scheduling
problem rather than specified only by the candidate count $N$. For
independent candidates, the main execution choice is how many candidates to
place in each generation call. A simple heuristic is to use the largest feasible batch size. Let
$b_{\max}$ be the largest number of candidates that fits the available GPU
memory and other system constraints. We select
\[
b = \min(N,b_{\max}), \qquad
C = \left\lceil \frac{N}{b} \right\rceil,
\]
where $b$ is the maximum number of candidates per call and $C$ is the number
of generation calls. If $N$ is not divisible by $b$, the final call contains
the remaining candidates. This policy minimizes the number of calls while
keeping the candidate budget fixed. For example, if all eight candidates fit in memory, our results favor
$1\times8$. If only four candidates fit at once, $2\times4$ is preferred
over schedules with more, smaller calls. The same rule naturally extends to
larger candidate budgets.

\begin{table}[t]
\centering
\caption{Practical guidance for selecting a candidate-generation schedule.}
\label{tab:scheduling-guidance}
\footnotesize
\renewcommand{\arraystretch}{1.08}
\begin{tabularx}{\columnwidth}{@{}XX@{}}
\toprule
Situation & Suggested approach \\
\midrule
All candidates fit in memory
& Use one generation call with batch size $N$ \\

Memory limits batch size
& Use the largest feasible batch and the fewest calls \\

Candidates depend on previous outputs
& Sequential calls may be required \\

Continuous-serving environment
& Coordinate with the serving scheduler \\

Additional latency or resource limits
& Choose a feasible schedule; among feasible options, prefer fewer calls \\
\bottomrule
\end{tabularx}
\end{table}
This heuristic is a starting point rather than a universal optimizer. The
best schedule may change with model size, sequence length, available memory,
concurrent traffic, and the inference engine. A more general scheduler could
predict the energy, latency, and memory cost of candidate schedules and select
the lowest-cost configuration that satisfies the system constraints.
\section{Limitations and Future Work}

Our experiments cover two models, two GPU architectures, and batch-scheduled
Hugging Face generation. The measured ratios should therefore not be assumed
to hold unchanged for larger models, other GPU generations, continuous
batching systems, or multi-GPU inference. The fixed-$N$ schedules use independently sampled candidates. Although their
candidate counts are identical and their logical generated-token volumes are
closely matched, we do not isolate the contribution of individual mechanisms
such as repeated call overhead, prompt processing, synchronization, or
batching effects. The results should therefore be interpreted as the
end-to-end cost of each generation schedule. Energy measurements represent gross GPU-device energy over the measured
query interval. They do not include whole-node energy and do not subtract
idle GPU power. Future work can extend the study to larger models, continuous batching,
quantization, speculative decoding, and multi-GPU execution. 


\section{Conclusion}

Candidate count $N$ tells us how many responses are generated, but not how
they are executed. Our experiments show that this execution choice can
substantially change the systems cost of multi-candidate inference. At fixed
$N=8$, splitting candidates across more generation calls consistently
increases latency, GPU-hours, and gross GPU-device energy while reducing
throughput. On A100 GPUs, $8\times1$ uses $4.64\times$ as much gross GPU-device energy
as $1\times8$ for Phi-3 and $4.86\times$ as much for Qwen. The same
scheduling pattern remains across independent A100 jobs and also appears in
the short-output SciQ/V100 setting. For independent candidates, our results support a practical rule: when
memory allows it, use fewer generation calls with larger batch sizes.
Multi-candidate evaluations should report generation schedule and systems
metrics alongside candidate count so that their cost can be reproduced and
compared.

\section*{Acknowledgment}

This project was supported by the National Science Foundation under grant
\#2211982. This work utilized the Nova high-performance computing cluster at
Iowa State University and the Delta system at the National Center for
Supercomputing Applications (NCSA) through allocation CIS240855. We
acknowledge computational support from the Iowa State University Research IT
Unit. Some of the Nova HPC equipment was purchased through funding provided
by the National Science Foundation under MRI grant 2018594. We also
acknowledge support for ACCESS through U.S. National Science Foundation
grants 2138259, 2138286, 2138307, 2137603, and 2138296.

\end{document}